\documentclass[letterpaper]{article} 
\usepackage[]{aaai23}  
\usepackage{times}  
\usepackage{helvet}  
\usepackage{courier}  
\usepackage[hyphens]{url}  
\usepackage{graphicx} 
\usepackage{natbib}  
\usepackage{caption} 
\usepackage{algorithm}
\usepackage[noend]{algpseudocode} 
\let\oldReturn\Return
\renewcommand{\Return}{\State\oldReturn}

\usepackage{newfloat}
\usepackage{listings}
\DeclareCaptionStyle{ruled}{labelfont=normalfont,labelsep=colon,strut=off} 
\floatstyle{ruled}
\newfloat{listing}{tb}{lst}{}
\floatname{listing}{Listing}
\usepackage[cmex10]{amsmath}
\usepackage{amsfonts}
\usepackage{amsthm}

\usepackage{multicol}

\usepackage{xspace}

\newcommand\moduleName[1]{\textsf{#1}\xspace}
\newcommand{\MAMO}{\moduleName{MAMO}}
\newcommand{\MAPF}{\moduleName{MAPF}}
\newcommand{\MfM}{\moduleName{\fontfamily{pbk}\selectfont M{\footnotesize 4}M}}
\newcommand{\EMfM}{\moduleName{\fontfamily{pcr}\selectfont E-\fontfamily{pbk}\selectfont M{\footnotesize 4}M}}

\newcommand{\KPIECE}{\textsc{KPIECE}\xspace}

\usepackage[table,xcdraw,dvipsnames,x11names]{xcolor}
\usepackage{color}

\colorlet{documentLinkColor}{red}
\colorlet{documentUrlColor}{blue}
\colorlet{documentCitationColor}{ForestGreen}

\usepackage{graphicx}

\newcommand{\X}{\mathcal{X}}

\newcommand{\R}{\mathcal{R}}
\newcommand{\Obs}{\mathcal{O}}
\newcommand{\OI}{\mathcal{I}}
\newcommand{\OM}{\mathcal{M}}

\usepackage{textcomp,gensymb}

\theoremstyle{proposition}

\usepackage{titlesec}

\usepackage[binary-units]{siunitx}
\usepackage{booktabs}
\usepackage{multirow}
\usepackage{cite}

\usepackage{mathtools}

\title{Planning for Manipulation among Movable Objects: Deciding Which Objects Go Where, in What Order, and How\footnote{This work was in part supported by ARL grant W911NF-18-2-0218 and ONR grant N00014-18-1-2775.}}

\author{
    Dhruv Saxena,
    Maxim Likhachev
}
\affiliations{
    Robotics Institute, Carnegie Mellon University\\
    \{dsaxena, mlikhach\}@andrew.cmu.edu
}
\usepackage{bibentry}

\begin{document}

\maketitle

\begin{abstract}

We are interested in pick-and-place style robot manipulation tasks in cluttered and confined 3D workspaces among movable objects that may be rearranged by the robot and may slide, tilt, lean or topple. 
A recently proposed algorithm, \MfM, determines \emph{which} objects need to be moved and \emph{where} by solving a Multi-Agent Pathfinding  \raisebox{1pt}{(}\MAPF{\raisebox{1pt}{)}} abstraction of this problem.
It then utilises a nonprehensile push planner to compute actions for \emph{how} the robot might realise these rearrangements and a rigid body physics simulator to check whether the actions satisfy physics constraints encoded in the problem.
However, \MfM greedily commits to valid pushes found during planning, and does not reason about orderings over pushes if multiple objects need to be rearranged.
Furthermore, \MfM does not reason about other possible \MAPF solutions that lead to different rearrangements and pushes.
In this paper, we extend \MfM and present Enhanced-\MfM (\EMfM) -- a systematic graph search-based solver that searches over orderings of pushes for movable objects that need to be rearranged and different possible rearrangements of the scene.
We introduce several algorithmic optimisations to circumvent the increased computational complexity, discuss the space of problems solvable by \EMfM and show that experimentally, both on the real robot and in simulation, it significantly outperforms the original \MfM algorithm, as well as other state-of-the-art alternatives when dealing with complex scenes.

\end{abstract}


\section{Introduction}
\begin{figure}[t]
    \centering
    \includegraphics[width=\columnwidth]{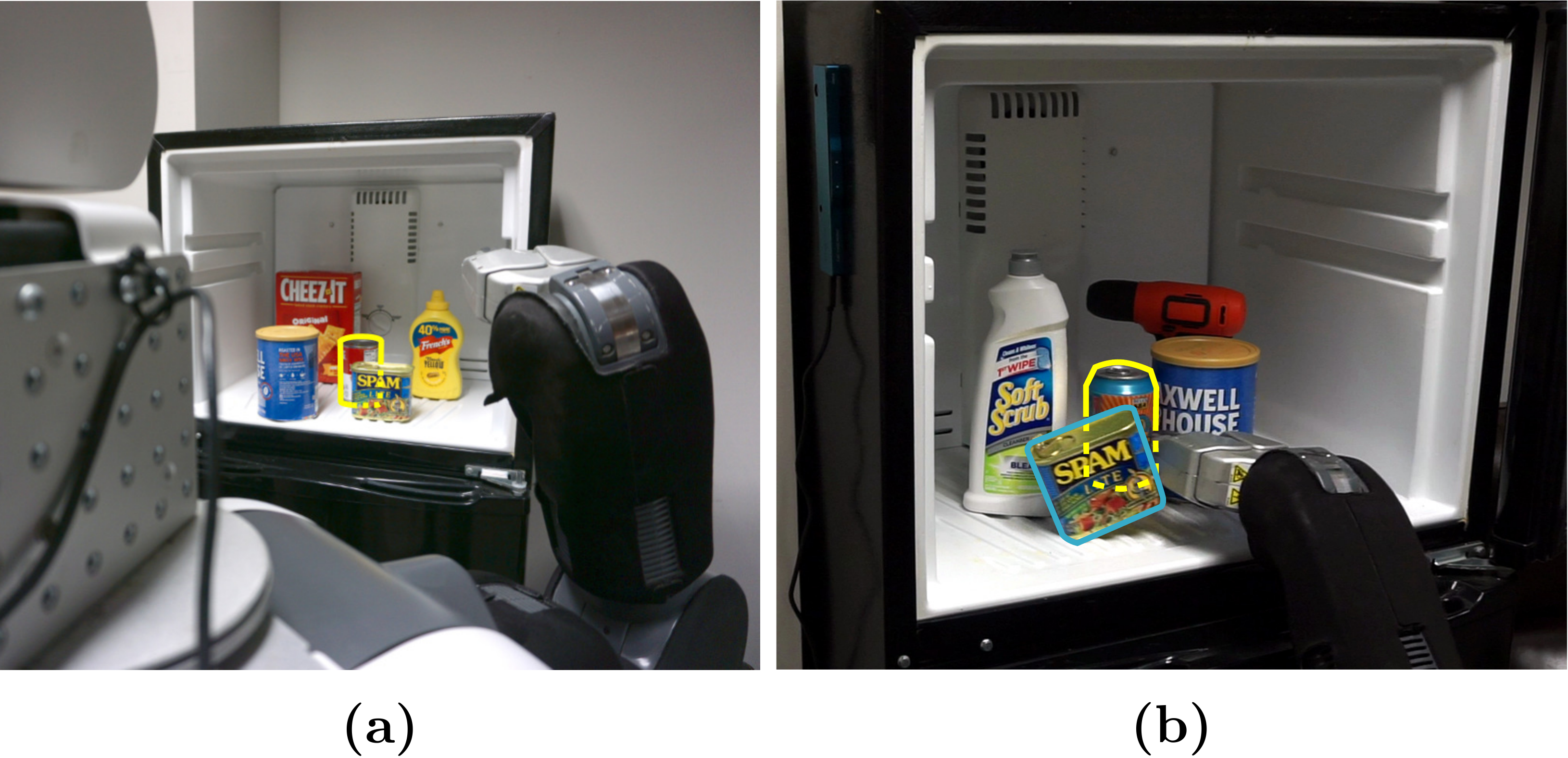}
    \caption{(a)~The tomato soup can (yellow outline) is the object-of-interest (OoI) to be retrieved. The potted meat can and coffee can in front of it must be rearranged out of the way in order to retrieve the OoI and solve the \MAMO problem. (b)~Trying to retrieve the beer can (OoI, yellow outline) leads to a complex interaction with the movable potted meat can being tilted by the robot arm.}
    \label{fig:intro_fridge}
\end{figure}

Simple pick-and-place robot manipulation tasks can be difficult to solve for motion planning algorithms that do not reason about how `movable' objects in the confined workspace might need to be rearranged in order to find a feasible solution path.
Such situations are commonly encountered when robot arms have to grasp and extract desired objects from cluttered shelves or pack several objects in a box.
Solving these ``Manipulation Among Movable Objects'' \raisebox{0.5pt}{(}\MAMO{\raisebox{0.5pt}{)}} problems~\cite{alami,StilmanMAMO} requires a planning algorithm to decide \emph{which} objects should be moved~\cite{Hauser14}, \emph{where} to move them, and \emph{how} they may be moved.
For the scene shown in Figure~\ref{fig:intro_fridge} (a), the tomato soup can is the ``object-of-interest'' (OoI) to be retrieved.
In order to do so, the PR2 robot must first move the coffee can and potted meat can out of the way so that the grasp pose for the OoI becomes reachable.

Existing state-of-the-art approaches in literature commonly assume prehensile (pick-and-place) rearrangement actions, e.g.~\citet{StilmanMAMO,wang2022lazy} and/or planar robot-object and object-object interactions, e.g.~\citet{BergSKLM08,homology_nonprehensile}.
Prehensile rearrangements not only preclude manipulation of big, bulky and otherwise ungraspable objects, but also assume access to known grasp poses for all movable objects and availability of stable placement locations for them in a cluttered and confined workspace.
The planar world assumption does not account for realistic physics interactions between objects in a real-world scene.
In contrast to these, our emphasis is on solving \MAMO problems (i) in a 3D workspace where robot actions can lead to complex multi-body interactions where objects tilt, lean on each other, slide, and topple (Figure~\ref{fig:intro_fridge} (b)); and (ii) with nonprehensile push actions for rearranging the clutter in the scene.
With this we allow for more seamless and natural manipulation that rearranges objects aside without picking them up while considering complicated toppling, sliding, and leaning effects.

The \MAMO problem definition includes information about which objects are \emph{movable} and which are static or \emph{immovable} obstacles.
All objects have a set of \emph{interaction constraints} associated with them that define valid robot-object and object-object interactions in the workspace.
Interaction constraints encode that neither the robot nor any other object can make contact with immovable obstacles (an object that cannot be interacted with, such as a wall), and movable objects cannot fall off the shelf, tilt too far (beyond $\SI{25}{\degree}$), or move with a high instantaneous velocity (above $\SI{1}{\meter\per\second}$).
These constraints help model realistic and desirable robot-object interactions since we want to prevent robots from carelessly hitting, pushing or throwing objects around.
In order to forward simulate the effect of nonprehensile robot pushes on the objects, we use a rigid body physics simulator to evaluate the interaction constraints and determine the resultant state of the workspace.

In recent work~\cite{Saxena23}, we proposed the \MfM algorithm for \MAMO to answer the questions of \emph{which} objects to move \emph{where}, and \emph{how}.
\MfM (``Multi-Agent Pathfinding for Manipulation Among Movable Objects'') relies on an \MAPF abstraction of \MAMO problems where the movable objects are artificially actuated agents with the goal of avoiding collisions with (i) the robot arm as it retrieves the OoI, (ii) each other, and (iii) immovable obstacles.
A solution to this \MAPF abstraction 
informs \MfM of \emph{which} objects must be moved and \emph{where} so that the OoI can be retrieved to solve the \MAMO problem.
\MfM then samples nonprehensile pushes to try and realise the rearrangements suggested by the \MAPF solution in the real-world, thereby addressing the third question of \emph{how} objects may be moved.

However, \MfM is greedy and can fail to find solutions in many cases where one may exist.
It is greedy in three different ways.
First, \MfM does not search over all possible orderings of object rearrangements if multiple movable objects need to be moved.
It greedily commits to the first valid push it finds and continues searching for a solution from the resultant state of that push.
Second, \MfM never reconsiders solving the \MAPF problem again for a different solution that might require objects to be rearranged differently.
As such, it does not search over all possible rearrangements of the scene.
This is important because in cases where an object cannot be rearranged successfuly as per the \MAPF solution (perhaps due to robot kinematic limits, the presence of immovable obstacles, interaction constraint violations etc.), we must replan the \MAPF solution and consider a different way to rearrange the scene that may indeed be feasible.
Finally, even in cases where we successfully rearrange an object to a particular location, \MfM never reconsiders moving that object differently, which may be required if no solution can be found from the resultant state of the valid push.

This paper extends \MfM and presents Enhanced-\MfM (\EMfM), an algorithm that addresses all three shortcomings of \MfM and does so by searching a much larger space for solutions.
It considers different orderings for rearranging objects, replans \MAPF solutions as and when required, and considers different ways to rearrange any particular object.
Although the search space for \EMfM grows tremendously as a result, \EMfM exploits the information gained during its execution to reduce redundant exploration of the solution space.
There is redundancy in considering the same or similar pushing actions for an object in different nodes of \EMfM search tree such that if one action succeeds or fails (i.e. its validity is determined by forward simulating it for interaction constraint verification), it is likely that the other actions will succeed or fail as well.
We exploit this by introducing caching of positive and negative simulation results and learning a probabilistic estimate of solving a particular subtree of the search, and use these within \EMfM to bias its exploration.
We make the following contributions as part of our \EMfM algorithm:
\begin{enumerate}
    \item A best-first graph search for \MAMO problems that searches over orderings of object rearrangements, different rearrangements of the scene, and different ways to rearrange each object.
    \item Caching results of successful (valid) pushes to avoid simulations of similar pushes repeatedly.
    \item Caching results of unsuccessful (invalid) pushes to feedback information to the \MAPF solver to efficiently search the space of rearrangements of the scene.
    \item A learned probabilistic model for solving a particular subtree to bias exploration of the best-first search.
    \item Significant quantitative improvements over \MfM and several other state-of-the-art \MAMO baselines.
\end{enumerate}



\section{Related Work}


In recent years, \MAMO planning algorithms have continued to rely on at least one of two simplifying assumptions.
The first limits the action space of the robot to prehensile or pick-and-place rearrangements of movable objects~\cite{StilmanMAMO,KrontirisSDKB14,KrontirisB15,LeeCNPK19,NamLCCK20,ShomeB21,wang2022lazy}.
This simplifies the planning problem as grasped objects behave as rigid bodies attached to the robot end effector, and rearrangement paths can be computed by avoiding collisions with other objects in the scene.
It is important to note that these paths can only be found if we assume (i) all objects that may need to be rearranged are graspable by the robot, (ii) we have known grasp poses for all graspable objects, (iii) the existence of stable placement locations for these objects in the cluttered workspace, and (iv) a relatively large volume of object-free space so that collision- and contact-free rearrangement paths with a grasped object exist.
In our work, we make none of these assumptions, instead relying on nonprehensile pushing actions to rearrange the scene.
This allows us to manipulate a much larger set of objects, and also lets us rearrange multiple objects simultaneously.
However, using these actions within a planning algorithm necessitates the ability to accurately predict their effect on the configurations of movable objects in order to compute the resultant state of the world after the push.

The second simplification assumes planar robot-object and object-object interactions, while allowing nonprehensile push actions for rearrangement.
This planar assumption halves the size of the configuration space of movable objects from $SE(3)$ to $SE(2)$.
To predict the effect of push actions on the scene, some existing algorithms make use of simple analytical or learned physics models~\cite{BergSKLM08,DogarS12,HuangHYB22}, while others use computationally cheap 2D physics simulators~\cite{King-2016-104000,Huang}.
Assuming planar interactions does not capture the complex multi-body physics of the 3D real-world where objects may tilt, lean on each other, topple etc., something we account for in the \EMfM algorithm.
As part of our experiments, we compare against an implementation of the algorithm from~\cite{DogarS12} that uses the same nonprehensile push planner as \EMfM in conjunction with a 3D rigid body physics simulator.
The original algorithm is not viable for our \MAMO problems as it uses a 2D analytical model to predict the result of planar robot-object interactions, and only allows the robot to rearrange one object at a time.

Existing work which uses a full 3D rigid body physics simulator to forward simulate the effect of robot actions on the scene does not account for the difficult interaction constraints we include in our \MAMO problems which makes it harder to find a feasible solution.
Instead, they either only deal with simple constraints~\cite{SelSim,SPAMP} where objects are not allowed to fall off the workspace shelf, or do not include any such constraints~\cite{gp_nonprehensile,homology_nonprehensile} and ignore cases where objects topple.
We include a comparison against~\cite{SelSim} in our experiments, albeit with the full interaction constraint set that \EMfM considers.

We also include comparisons against two general purpose sampling-based planning algorithms, \KPIECE~\cite{KPIECE} and RRT~\cite{RRT}, and our own recent \MfM algorithm developed for this \MAMO domain.
\KPIECE is a randomised algorithm developed for planning problems where it is expensive to determine the resultant state of an action (like querying a physics-based simulator for the effect of robot pushes).
\MfM, as discussed earlier, decouples the search for a solution to the \MAMO problem between solving an abstract \MAPF problem that reasons about the configuration of movable objects but does not require forward simulating a simulation-based model, and solving a simulation-based arm motion planning problem that does not need to search over the possible configurations of movable objects.


\section{Problem Formulation}

We are interested in solving \MAMO problems with a $q$ degrees-of-freedom robot manipulator $\R$ whose configuration space $\X_\R \subset \mathbb{R}^q$.
The workspace is populated with objects $\Obs = \{O_1, \ldots, O_n\}$ whose configuration spaces $\X_{O_k} \equiv SE(3)$.
We assume we know which objects $\OM \subset \Obs$ are \emph{movable} and which objects $\OI \subset \Obs$ are \emph{immovable}.
Each object $O_k$ is associated with \emph{interaction constraints} described earlier that help determine whether any state $x$ in the search space $\X \coloneqq \X_\R \times \X_{O_1} \times \cdots \times \X_{O_n}$ is valid or not.
The planning algorithm is provided the initial configurations of all movable objects (denoted as $\OM^{\text{init}}$) and immovable obstacles ($\OI$), information about which object is the ``object-of-interest'' (OoI), desired grasp pose for the OoI, and a ``home'' configuration outside the workspace shelf where the OoI must be moved.
Our goal is to find a path of valid states in $\X$ that successfully retrieves the OoI from the cluttered and confined workspace shelf.


We make no assumptions about the \MAMO problem being \emph{monotone} where each movable object may only be moved once, and we allow the robot to rearrange several movable objects at the same time.
We do assume access to a rigid body physics simulator to evaluate the effect of robot actions on the states of the objects in the workspace.


\section{\EMfM}

This paper presents the \EMfM algorithm, an enhanced version of our previous \MfM algorithm.
In this section we provide details about \EMfM, the \MAPF abstraction and nonprehensile push planner used within it, and discuss when and why \EMfM will solve a \MAMO problem (or not).

In order to solve the \MAMO problems of interest to us, \EMfM must answer questions about \emph{which} objects should be moved, \emph{where} they may be moved, and \emph{how} the robot can move them.
Like \MfM, it relies on two modules to answer these questions -- an \MAPF solver~\cite{SharonSFS15} is used to answer the first two questions, while our nonprehensile push planner uses the \MAPF solution to try and answer the third.
Unlike \MfM however, \EMfM runs a best-first search over a graph $G = (V, E)$ with the help of these two modules.
The vertices $v \in V$ represent a set of configurations (alternatively a \emph{rearrangement}) of the movable objects $\OM$ in the scene.
Edges $e = (u, v) \in E$ represent a successful rearrangement action changing the configuration of \emph{at least} one $O_m \in \OM$ between $u$ and $v$.
The overall \EMfM search expands vertices in an order dictated by some priority function $f : V \rightarrow \mathbb{R}_{\geq 0}$.
In contrast \MfM (i) greedily commits to the first valid push found (it does not search over orderings of rearrangements of multiple movable objects like \EMfM), (ii) only obtains a single \MAPF solution for each rearrangement it sees (it never replans the \MAPF solution based on the result of push actions like \EMfM), and consequently (iii) only tries to rearrange a movable object along a single \MAPF solution path for each rearrangement (it does not consider alternate ways to push an object for the same rearrangement like \EMfM).


\subsection{Main Algorithm}

Algorithm~\ref{alg:emfm} contains the pseudocode for \EMfM.
Initially, \EMfM computes a trajectory $\gamma_\text{OoI} \subset \X_\R$ for the robot to grasp and extract the OoI while pretending no movable objects $\OM$ exist in the scene (Line~\ref{line:first_traj}).
The argument for the $\textsc{PlanRetrieval}$ function is the set of objects to be considered as obstacles during planning.
The volume occupied by the robot arm and OoI during execution of $\gamma_\text{OoI}$, written as $\mathcal{V}(\gamma_\text{OoI})$, creates a \emph{negative goal region} (NGR)~\cite{DogarS12}.
We define an NGR parameterised with a robot trajectory as some volume in the workspace that, if free of all objects, will lead to successful retrieval of the OoI upon execution of that robot trajectory.
Note that if the trajectory $\gamma_\text{OoI}$ cannot be found, the overall \MAMO problem as specified is unsolvable.
It may be solvable given a different grasp pose for the OoI, however grasp planning is beyond the scope of this work.
Once the initial NGR $\mathcal{V}(\gamma_\text{OoI})$ has been computed (Line~\ref{line:ngr}), \EMfM executes a best-first search using a priority queue ordered by $f$.

\begin{algorithm}[!t]
\begin{small}
\caption{\EMfM{}}\label{alg:emfm}
\begin{algorithmic}[1]
\Procedure{CreateVertex}{$\OM, v, \gamma$}
    \State $v^\prime.\OM \gets \OM, v^\prime.parent \gets v, v^\prime.\gamma \gets \gamma$
    \Return $v^\prime$
\EndProcedure

\Procedure{Done}{$v$}
    \If{$v.\OM \cap \mathcal{V}(\gamma_\text{OoI}) = \emptyset$}
        \Return true \label{line:mapf_done}
    \EndIf
    \State $\hat{\gamma}_\text{OoI} \gets \textsc{PlanRetrieval}(v.\OM \cup \OI)$
    \If{$\hat{\gamma}_\text{OoI}$ exists}
        \State $\gamma_\text{OoI} \gets \hat{\gamma}_\text{OoI}$
        \Return true \label{line:new_retrieval}
    \EndIf
    \Return false
\EndProcedure


\Procedure{ExpandState}{$v$}
    \State $\kappa \gets \textsc{InvalidGoals}(v)$ \label{line:invalid_goals}
    \State $v.\{\pi_m\}_{m \in \OM} \gets \textsc{Run}\MAPF(v, \kappa, \mathcal{V}(\gamma_\text{OoI}))$ \label{line:call_mapf}
    \If{\MAPF failed}
        \State Remove $v$ from $OPEN$ \label{line:close_vertex}
        \Return
    \EndIf
    \For{$m \in v.\OM$} \label{line:get_succs_begin}
        \If{$v.\pi_m \neq \emptyset$}
            \State $\gamma_m \gets \textsc{PlanPush}(v.\pi_m, v.\OM)$ \label{line:plan_push}
            \State $\OM^\prime, \texttt{valid} \gets \textsc{IsValid}(\gamma_m)$ \label{line:validate_push}
            \If{\texttt{valid}}
                \State $v^\prime \gets \textsc{CreateVertex}(\OM^\prime, v, \gamma_m)$ \label{line:create_succ}
                \State Insert $v^\prime$ into $OPEN$ with priority $f(v^\prime)$ \label{line:insert_succ}
            \Else
                \State Add final state in $v.\pi_m$ to $\textsc{InvalidGoals}(v)$ \label{line:add_invalid_goal}
            \EndIf
        \EndIf
    \EndFor
\EndProcedure

\Procedure{Main}{$\OM^{\text{init}}, \OI$}
    \State $\gamma_\text{OoI} \gets \textsc{PlanRetrieval}(\OI)$ \label{line:first_traj}
    \State Compute $\mathcal{V}(\gamma_\text{OoI})$ \label{line:ngr}
    \State $OPEN \gets \emptyset$, $v_\text{start} \gets \textsc{CreateVertex}(\OM^{\text{init}}, \emptyset, \emptyset)$
    \State Insert $v_\text{start}$ into $OPEN$ with priority $f(v_\text{start})$
    \While{$OPEN$ is not empty \textbf{and} time remains}
        \State $v \gets OPEN.\textsc{top}()$
        \If{\textsc{Done}$(v)$}
            \Return $\textsc{ExtractRearrangements}(v)$
        \EndIf
        \State $\textsc{ExpandState}(v)$ \label{line:expand_state}
    \EndWhile
    \Return $\emptyset$
\EndProcedure

\end{algorithmic}
\end{small}
\end{algorithm}

Every time a vertex $v$ is expanded from this queue during the search (Line~\ref{line:expand_state}), \EMfM calls an \MAPF solver (Line~\ref{line:call_mapf}). 
The set of solution paths returned by the \MAPF solver is then used by our nonprehensile push planner to generate and evaluate successor rearrangement states (the loop from Line~\ref{line:get_succs_begin}).
For each object $O_m$ that ``moves'' in the \MAPF solution, our push planner tries to compute a trajectory $\gamma_m \subset \X_\R$ to push $O_m$ along its \MAPF solution path $\pi_m$ (Line~\ref{line:plan_push}).
If $\gamma_m$ is found, it is forward simulated in a rigid body physics simulator for interaction constraint verification (Line~\ref{line:validate_push}).
If $\gamma_m$ successfully rearranges the scene, i.e. at least one object is moved and no constraints are violated, \EMfM generates a successor state $v^\prime$ with the resultant rearrangement and adds it to the queue (Lines~\ref{line:create_succ} and~\ref{line:insert_succ}).

A vertex $v$ is \emph{closed} in Line~\ref{line:close_vertex} and never re-expanded again if the \MAPF solver fails to return a solution in Line~\ref{line:call_mapf}.
Otherwise when a vertex $v$ is re-expanded, we ensure that the \MAPF solver returns a different solution than one obtained during any previous expansion of $v$ by including a set $\kappa$ of invalid goals (Line~\ref{line:invalid_goals}).
$\kappa$ contains configurations for each movable object $O_m \in v.\OM$ that \emph{cannot} be the final state in the path $\pi_m$ found by the \MAPF solver.
This helps \EMfM search over different \MAPF solutions for the same rearrangement $v.\OM$, thereby helping it search over different ways to rearrange $v.\OM$.

\EMfM terminates with $v$ as the goal state when the OoI can be successfully retrieved given the rearrangement $v.\OM$.
This may be achieved in one of two ways.
If the movable objects $\OM$ in $v$ have been successfully rearranged to be outside the initial NGR $\mathcal{V}(\gamma_\text{OoI})$, we know the robot can execute $\gamma_\text{OoI}$ to retrieve the OoI (Line~\ref{line:mapf_done}).
Alternatively, if a different trajectory $\hat{\gamma}_\text{OoI}$ (and therefore its NGR $\mathcal{V}(\hat{\gamma}_\text{OoI})$) can be found in the presence of all objects (movable and immovable) as obstacles, the robot can execute $\hat{\gamma}_\text{OoI}$ to retrieve the OoI without making contact with any other object (Line~\ref{line:new_retrieval}).


\subsection{\MAPF Abstraction}

\begin{figure}[t]
    \centering
    \includegraphics[width=\columnwidth]{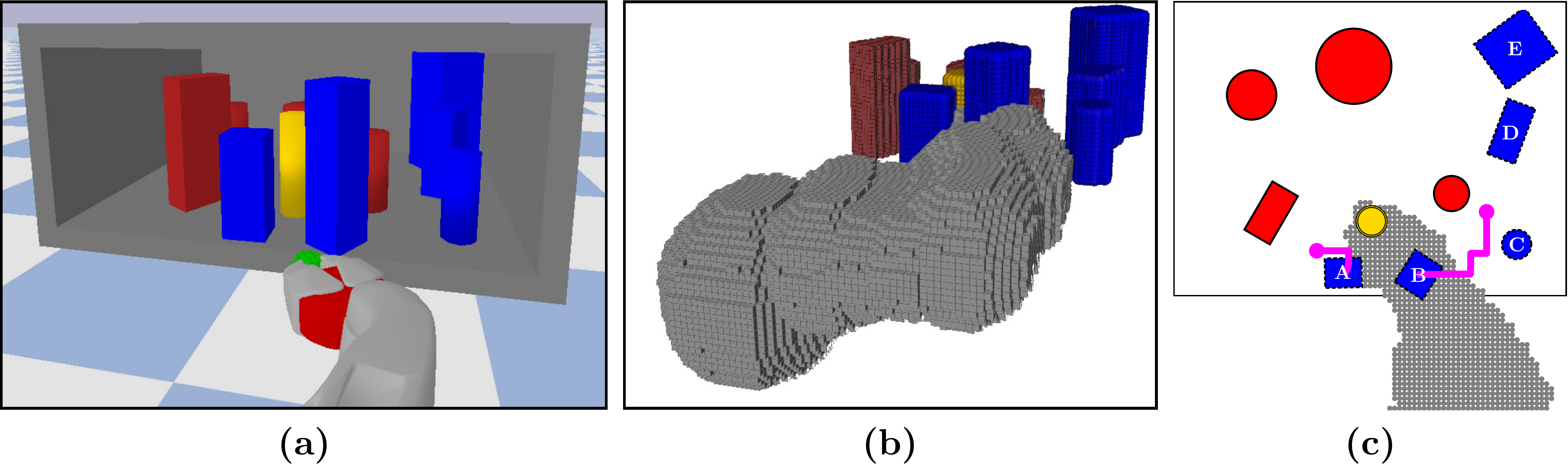}
    \caption{(a)~A \MAMO problem with five movable objects and four immovable obstacles, (b)~the initial NGR $\mathcal{V}(\gamma_\text{OoI})$ found for this scene (in gray), and (c)~a 2D projection of the scene with the \MAPF solution paths in pink. This \MAPF solution suggests that objects $A$ and $B$ should be rearranged as per the pink paths to be outside $\mathcal{V}(\gamma_\text{OoI})$.}
    \label{fig:mapf_solution}
\end{figure}

We make the observation that by solving a carefully constructed \MAPF problem with movable objects as agents, we can obtain information about which objects our \MAMO planner should consider rearranging and where.
We use Conflict-Based Search (CBS)~\cite{SharonSFS15} as the solver for our abstract \MAPF problem. 
For a vertex $v$ in \EMfM, we include all movable objects as agents in CBS starting at their current poses in $v.\OM$.
Each agent is assigned a goal of being outside the initial NGR $\mathcal{V}(\gamma_\text{OoI})$, while avoiding collisions with each other and all immovable obstacles $\OI$.
Although agent states in CBS specify their configuration in $\X_{O_m} \equiv SE(3)$, agents use a 2D action space on a four-connected grid in the \MAPF abstraction that only changes the $x-$ or $y-$coordinates of their state.
The CBS solution for this \MAPF problem is a set of paths $\{\pi_m\}_{m \in \OM}$ such that $O_m$ ends up outside $\mathcal{V}(\gamma_\text{OoI})$ after following $\pi_m$.
Figure~\ref{fig:mapf_solution} shows a simulated \MAMO problem, the initial NGR $\mathcal{V}(\gamma_\text{OoI})$ for the scene, and a 2D projection of the scene which shows the \MAPF solution found.

\begin{figure}[t]
    \centering
    \includegraphics[width=0.8\columnwidth]{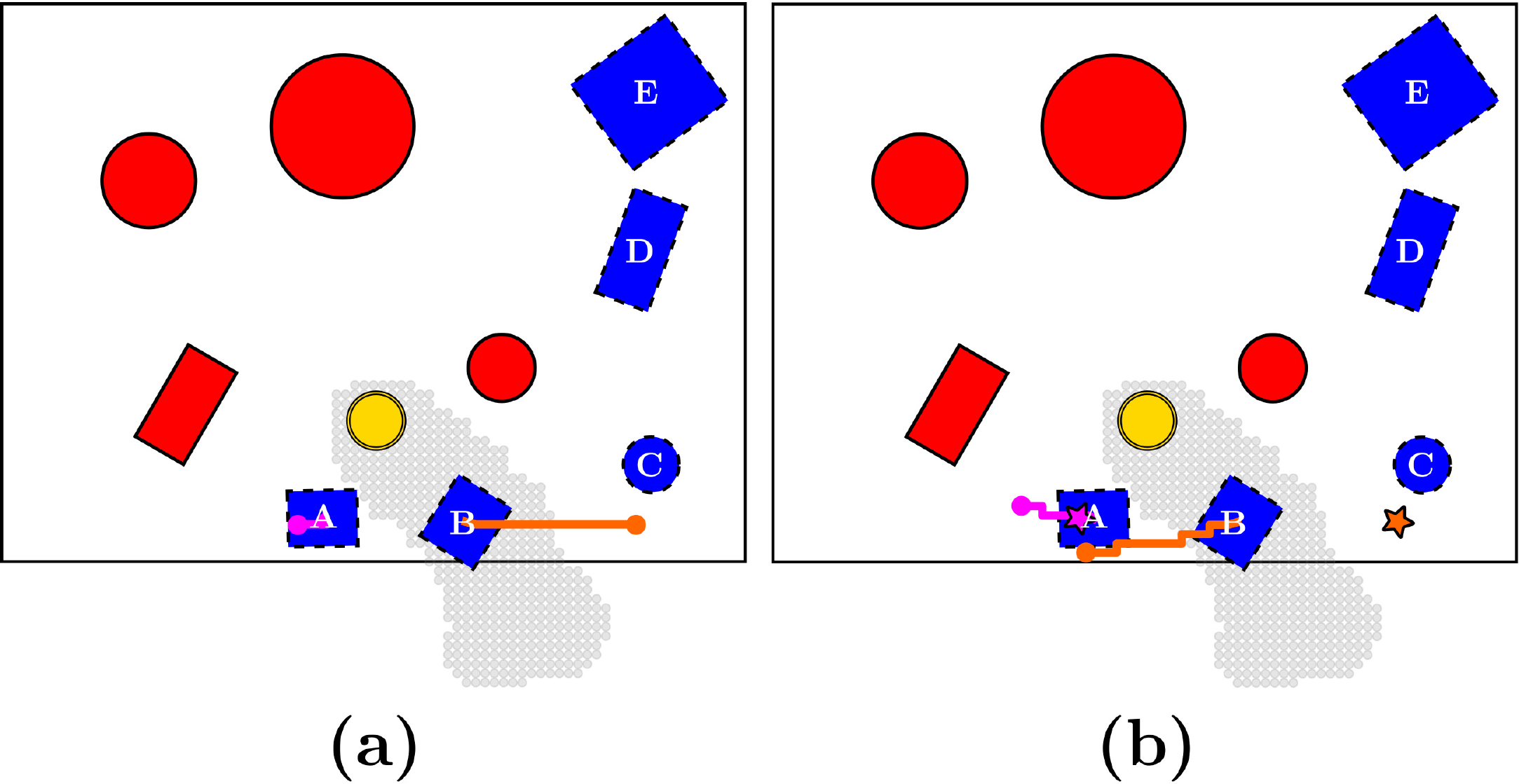}
    \caption{(a)~First \MAPF solution that led to invalid pushes $\gamma_A$ and $\gamma_B$, (b)~Adding the final states of $\pi_A$ and $\pi_B$ (colour coded stars) to $\kappa$ leads to a new \MAPF solution.}
    \label{fig:mapf_constraints}
\end{figure}

In order to search over all possible ways to rearrange $v.\OM$, \EMfM includes a set of invalid goals $\kappa$ when it calls CBS in Line~\ref{line:call_mapf}.
For each vertex $v$, $\kappa$ includes information about where each object $O_m \in v.\OM$ \emph{cannot} end up in any future CBS solution.
During a previous expansion of $v$, if path $\pi_m$ led to an invalid push $\gamma_m$, the final state of $\pi_m$ is added as an invalid goal for $O_m \in v.\OM$ since we failed to rearrange $O_m$ along $\pi_m$ (Line~\ref{line:add_invalid_goal}).
As a result, the next solution from CBS would return a new path $\pi^\prime_m \neq \pi_m$ which in turn would cause our push planner to consider a different rearrangement action.
When all invalid pushes from $v$ have been included in the previous call to CBS, we also include the final states of valid pushes found previously as invalid goals in $\kappa$ so as to ensure we consider all possible ways to rearrange $v.\OM$.
If CBS fails to find a solution, or if no new invalid goals can be added to $\kappa$ from the last call to CBS, we \emph{close} vertex $v$ and stop it from being re-expanded (Line~\ref{line:close_vertex}) since no new way to rearrange $v.\OM$ can be found.
Figure~\ref{fig:mapf_constraints} shows the effect of invalid goals $\kappa$ on the \MAPF solution -- when both pushes $\gamma_A$ and $\gamma_B$ computed as per the \MAPF solution in (a) failed, adding the final states of $\pi_A$ and $\pi_B$ to $\kappa$ results in a different \MAPF solution in (b).

\subsection{Nonprehensile Push Planner}

The goal for our push planner is to find robot trajectories $\gamma_m \subset \X_\R$ that rearrange a movable object along the path $\pi_m$ returned by our \MAPF solver.
If we are able to precisely rearrange each $O_m$ to the final state of $\pi_m$, we know the robot can execute $\gamma_\text{OoI}$ to solve the \MAMO problem.
In order to do so, we assume the push planner is provided a shortcut path $\pi_m$ (accounting for immovable obstacles $\OI$) in Line~\ref{line:plan_push}.

\begin{figure}[t]
    \centering
    \includegraphics[width=0.8\columnwidth]{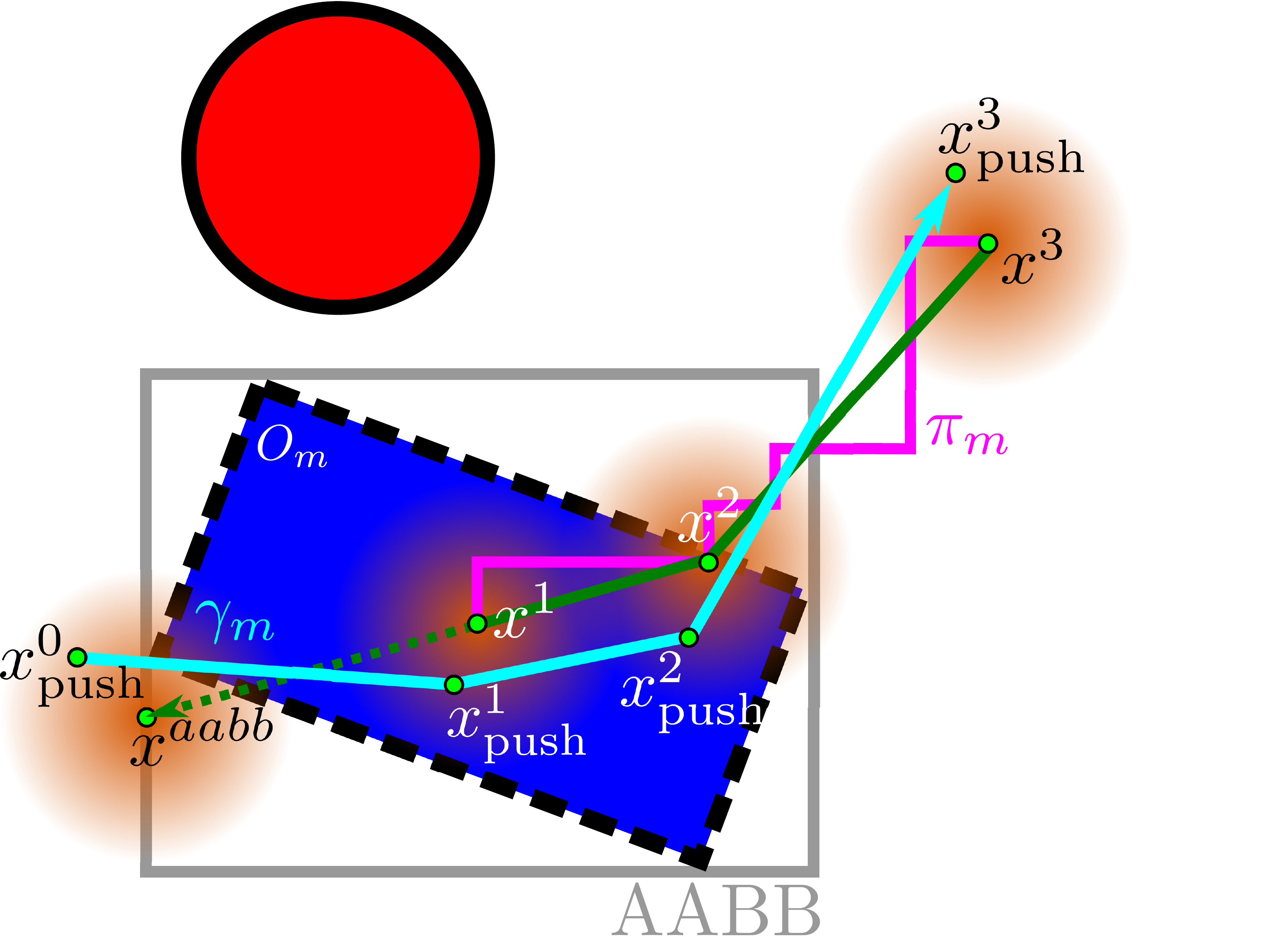}
    \caption{2D illustration of our push planner. Movable object $O_m$ is blue, and an immovable obstacle is drawn in red. The green path is obtained after shortcutting the pink \MAPF solution path $\pi_m$. $x^{aabb}$ is the point-of-intersection between the first segment $(x^1, x^2)$ and the axis-aligned bounding box for $O_m$. The cyan segments depict the path along which inverse kinematics is used to obtain $\gamma_m^i$.}
    \label{fig:push_planner}
\end{figure}

Each call to $\textsc{PlanPush}$ stochastically generates a robot trajectory $\gamma_m$, as shown in Figure~\ref{fig:push_planner}.
The starting location of the push, $x^0_{\text{push}}$, is sampled around the point of intersection $x^{aabb}$ between the first segment of $\pi_m$ and the axis-aligned bounding box of $O_m$.
If the push planner finds an approach path $\gamma_m^0 \subset \X_\R$ to $x^0_{\text{push}}$ in the presence of all objects (movable and immovable) as obstacles, it samples waypoints $x^i_{\text{push}}$ around the corresponding states $x^i$ in $\pi_m$.
It uses inverse kinematics to find segments of the push $\gamma_m^i$ between the waypoints $x^{i-1}_{\text{push}}$ and $x^i_{\text{push}}$.
If all $\gamma_m^i$ are found, the planner returns the final push trajectory as a concatenation of the individual pieces.
In this case, the push $\gamma_m$ is forward simulated in a rigid body physics simulator to detect if any interaction constraints are violated during its execution and get the resultant rearrangement of the scene.





\subsection{What \EMfM Can and Cannot Solve}

Solutions to \MAMO problems lie in a space $\X = \X_\R \times \X_{O_1} \times \cdots \times \X_{O_n}$ that grows exponentially with the number of movable objects.
There are several subtle reasons due to which \EMfM might fail to find solutions to complicated \MAMO problems in this space.
When trying to rearrange an object to a specific location, \EMfM only considers moving it along the particular path $\pi_m$ found by CBS and not along all such paths.
Its reliance on CBS and our push planner also means that \EMfM might fail to find interesting non-monotone solutions where it must rearrange an object $O_a$ partially along its solution path $\pi_a$ before moving $O_b$ along $\pi_b$ and finally going back to move $O_a$ the remainder of the way along $\pi_a$.
The \MAPF abstraction itself uses a simple 2D action space which fails to capture all possible rearrangements of the scene in $\X_{O_1} \times \cdots \times \X_{O_n}$.
Finally, \EMfM does not actively search over all OoI retrieval trajectories $\gamma_\text{OoI} \subset \X_\R$, and consequently the same NGR is used to specify goals for movable objects in all \MAPF calls.

Despite these limitations, by relying on our \MAPF abstraction and nonprehensile push planner, \EMfM does search over `allowed' (i) orderings of movable object rearrangements, (ii) potential rearrangements for the set of movable objects, and (iii) ways to rearrange the same movable object.
In doing so it makes progress towards a complete \MAMO planning algorithm that uses nonprehensile actions for rearrangement in a 3D workspace with complex multi-body interactions where movable objects may tilt, lean, topple etc.
Our quantitative analysis shows that \EMfM performs better than many state-of-the-art planning algorithms for these \MAMO problems.



\section{Speeding up the Algorithm}

This section discusses three algorithmic optimisations we propose as part of \EMfM that significantly improve its quantitative performace.


\subsection{Caching Unsuccessful Push Actions}

The goal set for movable objects in the \MAPF abstraction includes any configuration outside the initial NGR $\mathcal{V}(\gamma_\text{OoI})$ that is free of collision from all other objects.
Given vertex $v$, consider an object $O_m$ in $v.\OM$ and its corresponding path $\pi_m$ which achieves this goal.
If the resulting push $\gamma_m$ is invalid, the next call to CBS from $v$ will lead to a path $\pi_m^\prime \neq \pi_m$.
However, naively including the last state in $\pi_m$ as an invalid goal state for CBS will likely lead to the new path $\pi_m^\prime$ ending in a neighbouring state of the invalid goal (since CBS is an optimal \MAPF solver).
This in turn will lead to a push $\gamma_m^\prime \approx \gamma_m$ that is also likely to be invalid.

\begin{figure}[t]
    \centering
    \includegraphics[width=0.8\columnwidth]{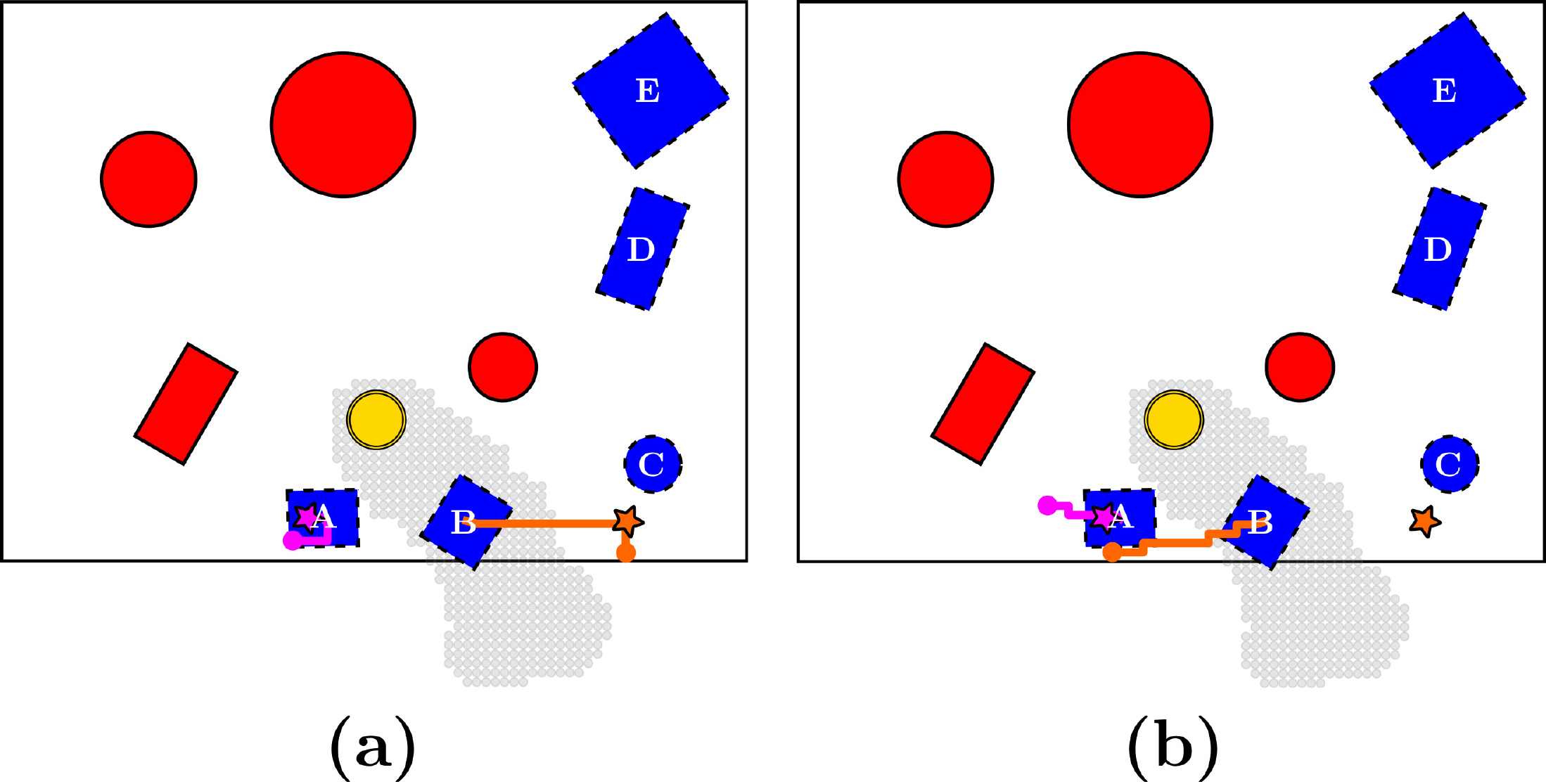}
    \caption{(a)~Only adding the final states of paths that led to invalid pushes as invalid goals results in a new \MAPF solution very similar to the previous one (from Figure~\ref{fig:mapf_constraints} (a)), (b)~Using our proposed approach that penalises goal states near known invalid goals, \EMfM is able to find diverse \MAPF solutions quicker.}
    \label{fig:neg_db_effect}
\end{figure}

To mitigate this, for every CBS call from a vertex $v$, for each object $O_m$, \EMfM caches the goals that were previously determined to be invalid in a nearest neighbour data structure.
We use this cached information to bias the solutions produced by CBS to avoid moving objects to states close to known invalid goals for the respective objects.
This biasing is done by assigning penalties to each potential goal location for each object $O_m$, and then during each low-level search within CBS finding the solution that minimises the summation of getting to a goal plus the penalty associated with the goal.
This can be done by introducing one single `pseudogoal' that the search searches towards and connecting all the potential goal locations to this `pseudogoal' with edges whose cost is proportional to the respective penalty.
This helps penalise paths to states close to known invalid goals, and lets \EMfM search the space of allowed rearrangements of $v.\OM$ more efficiently.
Figure~\ref{fig:neg_db_effect} (a) shows the new \MAPF solution when we include invalid goals naively -- the new paths $\pi_A^\prime$ and $\pi_B^\prime$ are very similar to $\pi_A$ and $\pi_B$ (from Figure~\ref{fig:mapf_constraints} (a)) and end in final states very close to the invalidated goals.
However, using the above explained approach that modifies the single-agent search to use our nearest neighbour data structure for invalid goals, we get a very different \MAPF solution in Figure~\ref{fig:neg_db_effect} (b).

\subsection{Caching Successful Push Actions}

During the search, CBS solutions in different vertices of the graph \EMfM may contain the same path $\pi_m$ for object $O_m$.
In such cases, if we have computed the push $\gamma_m$ in one of these vertices and found it to be valid in simulation while generating the successor rearrangement state, we would like to reuse this result whenever possible since simulating a push action is computationally expensive (around $2-6\SI{}{\second}$ per action simulation).
This reuse of successful push actions is enabled by storing the successful pushes in a database.

If a push $\gamma_m$ from path $\pi_m$ for object $O_m$ rearranged the scene from $v.\OM$ to $v^\prime.\OM$, we index into this database with the key $(O_m, \pi_m)$.
While any push is simulated, we keep track of objects that are $relevant$ for that push -- these are all objects whose configurations are changed between $v.\OM$ and $v^\prime.\OM$.
For each $(O_m, \pi_m)$ tuple, the database stores the value $(v.\OM, v^\prime.\OM, \gamma_m, relevant \text{ objects})$.
During the expansion of some other vertex $u$, if CBS returns the same path $\pi_m$ for $O_m$, we try to reuse the result of the stored push $\gamma_m$ to generate the successor state $u^\prime$ corresponding to rearranging $O_m$.
However, this reuse is only possible if all $relevant$ objects are in the same configurations in $v.\OM$ and $u.\OM$, and all other `irrelevant' objects in $u.\OM$ are in configurations that will not be affected by $\gamma_m$.
If both these conditions are true, we can simply reuse the result of $\gamma_m$ stored in the database to say that the $relevant$ objects in $u.\OM$ will be rearranged to their repective configurations $v^\prime.\OM$, and the `irrelevant' objects will remain unaffected.


\subsection{Learned Priority Function}

\EMfM searches an extremely large space for \MAMO solutions since it searches over orderings of object rearrangements, different rearrangements of the scene, and different ways to rearrange each object.
In an attempt to focus its search effort on more promising vertices of the search tree, we learn to predict the probability of a particular vertex leading to a solution for the \MAMO problem.
The learned function is used as the priority function $f$ in the best-first \EMfM search.
We predict the probability of a vertex $v$ leading to a solution based on features of the rearrangement $v.\OM$ -- the percentage volume of the initial NGR $\mathcal{V}(\gamma_\text{OoI})$ occupied by movable objects ($\phi_1$), the number of movable objects inside the NGR ($\phi_2$), and for each such object the product of its mass, coefficient of friction and percentage volume inside the NGR ($\phi_3$).
These features indicate how difficult it is to clear the NGR and therefore solve the \MAMO problem.
Our predictive model $f$ makes the Naive Bayes assumption~\cite{JohnL95} that these features are conditionally independent of the others.
Thus if $E$ is the event that vertex $v$ leads to a \MAMO solution,
\begin{align*}
    f(v) = P(E) \times P(\phi_1 \,\lvert\, E) &\times P(\phi_2 \,\lvert\, E) \\
    &\times \Pi_{O_m \in \mathcal{V}(\gamma_\text{OoI})} P(\phi_3 \,\lvert\, E)
\end{align*}

We model $P(\phi_1 \,\lvert\, E)$ as a beta distribution, and $P(\phi_2 \,\lvert\, E)$ and $P(\phi_3 \,\lvert\, E)$ are both exponential distributions.
Their parameters are estimated via maximum likelihood estimation from a dataset of self-supervised \MAMO problems.
We generate this set of problems by running \EMfM breadth-first on 20 \MAMO scenes with a $\SI{10}{\minute}$ planning timeout.
We store each vertex generated during these \EMfM runs as a separate \MAMO problem, and then run \EMfM depth-first with a $\SI{60}{\second}$ timeout on these problems to get our training dataset of $2400$ datapoints.


\section{Experimental Analysis}

This section compares the performance of \EMfM against several \MAMO baselines, studies the effect of the algorithmic improvements we propose in an ablation study, and provides results from real-world experiments on a PR2 robot.


\subsection{Simulation Experiments Against \MAMO Baselines}

\begin{figure}[t]
    \centering
    \includegraphics[width=\columnwidth]{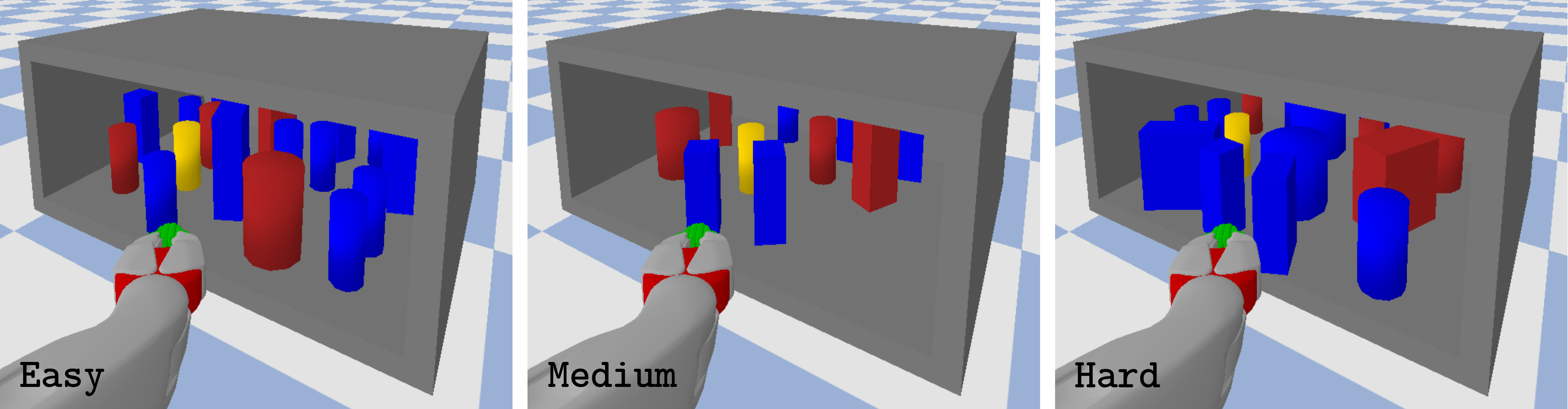}
    \caption{Example \texttt{Easy}, \texttt{Medium}, and \texttt{Hard} scenes.}
    \label{fig:problem_difficulty}
\end{figure}

Our simulation experiments randomly generate \MAMO workspaces with one OoI, four immovable obstacles, and five, ten or fifteen movable objects.
All object properties (shapes, sizes, mass, coefficient of frictions, initial poses) are randomised and known to each planner prior to planning.
We categorise these workspaces into three difficulty levels based on the number of movable objects inside the initial NGR $\mathcal{V}(\gamma_\text{OoI})$.
Problems are \texttt{Easy}, \texttt{Medium}, or \texttt{Hard} depending on whether there are one, two, or more than two movable objects overlapping with the initial NGR.
Figure~\ref{fig:problem_difficulty} shows sample \MAMO workspaces of each difficulty level.
We test the performance of all algorithms on 98 \texttt{Easy}, 63 \texttt{Medium}, and 39 \texttt{Hard} problems with a $\SI{5}{\minute}$ planning timeout.

In addition to (1)~\MfM, we compare the performace of \EMfM against four other baselines.
(2)~We reimplement \textsc{Dogar}~\cite{DogarS12} to use our push planner with a physics-based simulator.
It recursively searches backwards in time for objects that need to be rearranged outside the most recent NGR.
If it rearranges an object successfully, the volume spanned by the rearrangement trajectory is added to the previous NGR and the recursion continues.
However, \textsc{Dogar} only has information about which objects need to be rearranged but not where they should be moved.
Our implementation randomly samples points outside the latest NGR as goal locations for our push planner.
(3)~\textsc{SelSim}~\cite{SelSim} interleaves planning a trajectory while simulating interactions with `relevant' objects with tracking the found trajectory in the presence of all objects.
If tracking violates interaction constraints, the `culprit' object is identified and added to the set of relevant objects for the next iteration.
\textsc{SelSim} uses simple motion primitives that change only one of the $q$ degrees-of-freedom of the robot, which does not lead to meaningful robot-object interactions in this domain.
Finally, we compare against the standard OMPL~\cite{OMPL} implementations of general-purpose sampling-based planning algorithms (4)~\textsc{RRT}~\cite{RRT} and (5)~\textsc{KPIECE}~\cite{KPIECE}.

\begin{table}[]
\centering
\begingroup
\setlength{\tabcolsep}{2pt}
\footnotesize
\begin{tabular}{@{}ccccccc@{}}
\toprule
\multirow{2}{*}{\textbf{Difficulty}}& \multicolumn{6}{c}{\textbf{Planning Algorithm}} \\
\cmidrule{2-7}
& \EMfM & \MfM & \textsc{Dogar} & \textsc{SelSim} & \textsc{RRT} & \textsc{KPIECE}\\ \midrule
\texttt{Easy} (98) & 97 & 78 & 7 & 16 & 33 & 16 \\
\texttt{Medium} (63) & 45 & 25 & 0 & 8 & 7 & 1 \\
\texttt{Hard} (39) & 15 & 7 & 0 & 1 & 1 & 0 \\ \bottomrule
\end{tabular}
\endgroup
\caption{Number of problems solved by various \MAMO planning algorithms in simulation experiments}
\label{tab:sim_exps}
\end{table}

\begin{figure}[t]
    \centering
    \includegraphics[width=0.88\columnwidth]{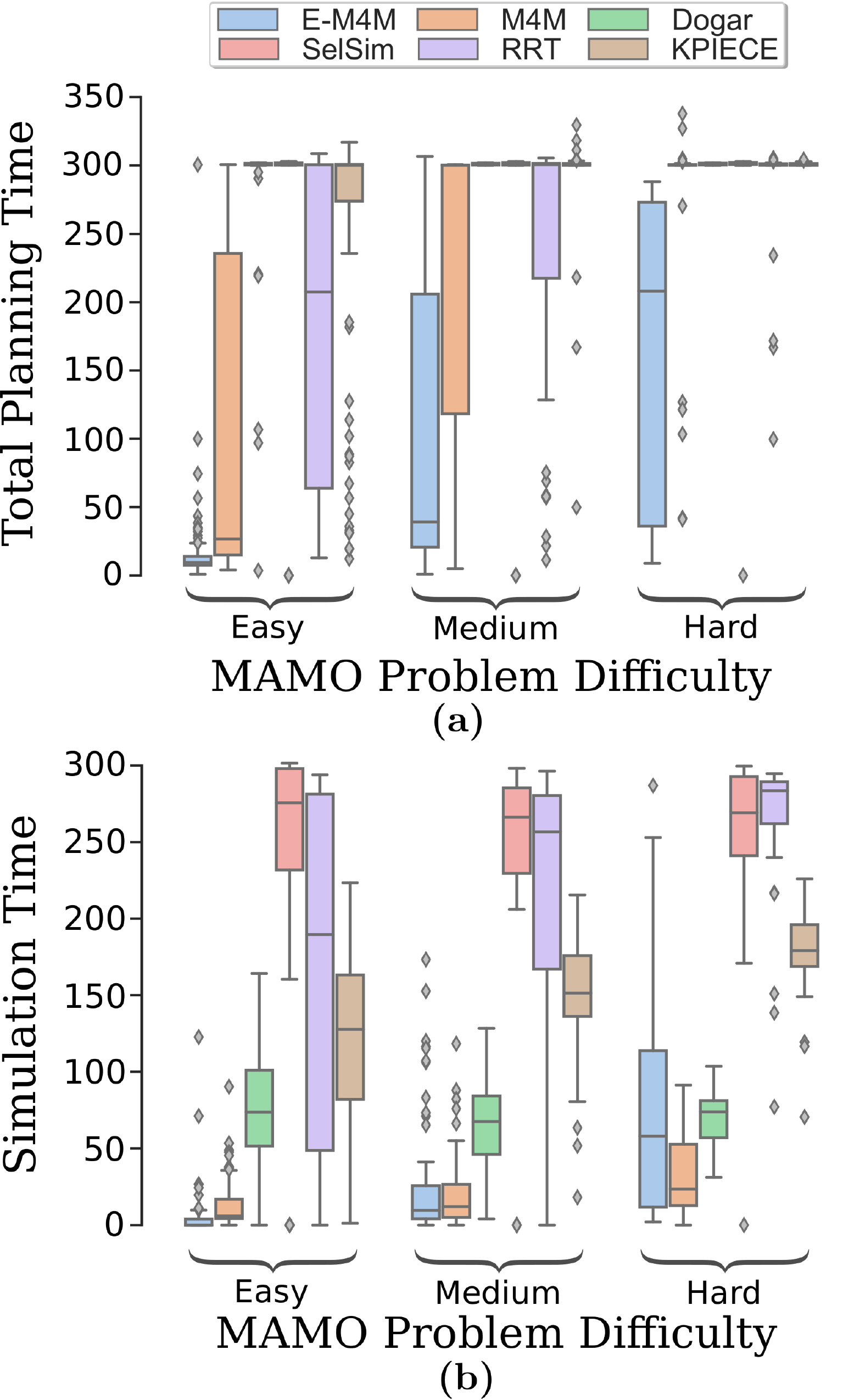}
    \caption{(a)~Total planning time and (b)~time spent querying a physics-based simulator for \MAMO planning algorithms across planning problems with varying difficulty levels.}
    \label{fig:baselines_results}
\end{figure}

Table~\ref{tab:sim_exps} contains the number of problems solved by each planning algorithm for the different difficulty levels.
It is apparent that \EMfM far outperforms all other algorithms, and that all baselines struggle to solve \texttt{Medium} and \texttt{Hard} problems.
The quantitative performance of all algorithms in terms of total planning time and time spent simulating robot actions is shown in Figure~\ref{fig:baselines_results}.
\EMfM is able to achieve a good balance of time spent computing robot actions (with the \MAPF solver and push planner) and forward simulating them for interaction constraint verification.
Since \MfM never replans the \MAPF solution, it spends most of its time trying to sample push actions to be simulated.
\textsc{Dogar} repeatedly fails to find solutions, even for simple problems, because it (i) has no information about where objects should move, choosing to randomly sample uninformed pushes instead, (ii) only considers pushes to be successful if they rearrange an object to be completely outside the NGR, and (iii) never considers rearranging an object more than once.
The performance of \textsc{SelSim} is particularly interesting.
It is only able to solve problems where the first planned path succeeds when tracked without any interaction constraints being violated, resulting in negligible planning times for its successes (and no time spent in simulation).
Otherwise, owing to its primitive action space, it spends most of its planning budget in simulation trying to rearrange the scene with small robot actions that are incapable of significantly changing object configurations.
\textsc{RRT} and \textsc{KPIECE} perform well as they are able to sample long robot motions that can rearrange objects with favourable physical properties (low masses and coefficients of friction) with high likelihood.

\subsection{\EMfM Ablation Study}

To study the effect of the algorithmic improvements we propose as part of \EMfM, we compare four different versions of \EMfM.
\textsc{Neg-DB} only caches information from unsuccessful push actions; 
\textsc{Pos-DB} only caches information from successful push actions; 
\textsc{No-DB} does not cache any information from push actions;
and \textsc{Tiebreak} assigns priorities by lexicographically tiebreaking \EMfM vertex feature vectors $(\phi_1, \phi_2, \sum_{O_m \in \mathcal{V}(\gamma_\text{OoI})} \phi_3)$.
\textsc{Neg-DB}, \textsc{Pos-DB}, and \textsc{No-DB} all use the learned priority function like \EMfM, while \textsc{Tiebreak} caches information from unsuccessful and successful push actions like \EMfM.

\begin{table}[]
\centering
\begingroup
\setlength{\tabcolsep}{2pt}
\footnotesize
\begin{tabular}{@{}cccccc@{}}
\toprule
\multirow{2}{*}{\textbf{Difficulty}}& \multicolumn{5}{c}{\textbf{Ablation}} \\
\cmidrule{2-6}
& \EMfM & \textsc{Neg-DB} & \textsc{Pos-DB} & \textsc{No-DB} & \textsc{Tiebreak}\\ \midrule
\texttt{Easy} (98) & 97 & 87 & 86 & 82 & 85 \\
\texttt{Medium} (63) & 45 & 24 & 29 & 25 & 36 \\
\texttt{Hard} (39) & 15 & 7 & 8 & 7 & 13 \\ \bottomrule
\end{tabular}
\endgroup
\caption{Number of problems solved \EMfM ablations}
\label{tab:ablation2}
\end{table}

\begin{figure}[t]
    \centering
    \includegraphics[width=0.9\columnwidth]{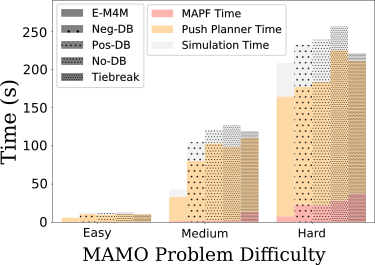}
    \caption{Median time spent calling the \MAPF solver, push planner, and simulator for different \EMfM ablations.}
    \label{fig:ablation_study}
\end{figure}

Table~\ref{tab:sim_exps} shows that each of these \EMfM ablations solve fewer \MAMO problems in comparison to \EMfM which combines all of them.
Quantitatively, their performance can be compared from the plots in Figure~\ref{fig:ablation_study}.
While there is no significant difference between the different ablations for \texttt{Easy} problems, for \texttt{Medium} and \texttt{Hard} problems we can see that performance degrades as we remove cached information.
\textsc{Pos-DB} performs worse than \textsc{Neg-DB} since it is not as likely for \EMfM to find the same push multiple times during a search as it is for it to require several different \MAPF solutions.
Finally, even with a naive tiebreaking based priority function, \textsc{Tiebreak} performs only slightly worse than \EMfM for \texttt{Hard} problems.
This suggests that the learned priority function (using the Naive Bayes assumption) is not as useful for these problems, perhaps due to the $\SI{60}{\second}$ timeout imposed during data collection being insufficient to result in a rich set of datapoints for \texttt{Hard} problems.

\subsection{Real-World Experiments}

We ran experiments with the PR2 robot with a refrigerator compartment as our \MAMO workspace (Figure~\ref{fig:intro_fridge}).
Problems were initialised with five objects from the YCB Object Dataset~\cite{YCB}.
The tomato soup can was always our OoI, while all other objects were initialised as movable.

We ran \EMfM on 20 perturbations of the scenes in Figure~\ref{fig:intro_fridge} with a $\SI{5}{\minute}$ planning timeout.
15 runs resulted in successful OoI retrieval, with the others failing due to unforeseen discrepancies between simulated and real-world robot-object interactions.
Four failures were due to inaccurate computation of coefficients of friction of movable objects.
One failure was the result of an object getting stuck in ridges in the real-world refrigerator shelf that were not modeled in simulation.
These discrepancies highlight the sim-to-real gap that \EMfM can suffer from, since it blindly relies on the result of the physics based simulator used in the algorithm.
On average, for the 15 successful retrievals, \EMfM took a total time of $39.3 \pm 28.2 \,\SI{}{\second}$ of which $0.9 \pm 1.2 \,\SI{}{\second}$ was spent calling the \MAPF solver, $33.5 \pm 25.8 \,\SI{}{\second}$ was spent planning pushes, and $7.1 \pm 5.3 \,\SI{}{\second}$ was spent simulating them.


\section{Conclusion and Future Work}

The Enhanced-\MfM algorithm presented in this paper builds upon our prior work on Multi-Agent Pathfinding for Manipulation Among Movable Objects~\cite{Saxena23}.
\EMfM utilises an \MAPF abstraction of \MAMO, a nonprehensile push planner, and a rigid body physics simulator within a best-first graph search for solving \MAMO problems that require determing \emph{which} movable objects should be moved, \emph{where} to move them, and \emph{how} they can be moved.
\EMfM searches over different orderings of object rearrangements, different rearrangements of the workspace, and different ways to rearrange the same object.

Currently, the \MAPF solver does not take into account any information about robot kinematics, movable object properties, or immovable obstacle poses (other than for collision checking against agents) when computing solution paths.
Since the \EMfM algorithm uses these paths downstream for nonprehensile push planning, in future work we wish to explore an experience-based learning formulation to take these factors into account as part of the \MAPF cost function to find paths more likely to result in valid pushes.

\bibliography{references}




\end{document}